\pdfoutput=1

\documentclass[11pt]{article}

\usepackage{emnlp2021} 

\usepackage{times}
\usepackage{latexsym}

\usepackage[T1]{fontenc}

\usepackage[utf8]{inputenc}

\usepackage{microtype}
\usepackage[ruled,vlined]{algorithm2e}

\newcommand{\eg}{e.g., }
\newcommand{\ie}{i.e., }

\newcommand{\secref}[1]{Section~\ref{#1}}
\newcommand{\equref}[1]{Eq.~(\ref{#1})}
\newcommand{\figref}[1]{Fig.~\ref{#1}}    

\usepackage{color}

\usepackage{amsmath}
\usepackage{amsfonts}
\usepackage{tikz}
\usepackage{mathrsfs}
\usepackage[ruled,vlined]{algorithm2e}
\usepackage[textsize=small]{todonotes}
\usepackage[inline,shortlabels]{enumitem}
\usepackage{multirow}
\usepackage{booktabs}
\usepackage{pifont}
\newcommand{\xmark}{\ding{55}}%

\title{A Simple Geometric Method for Cross-Lingual Linguistic Transformations with Pre-trained Autoencoders}

\author{Maarten De Raedt \\ Chatlayer.ai by Sinch \And  Fréderic Godin \\ Chatlayer.ai by Sinch \And Pieter Buteneers \\ Sinch \AND Chris Develder  \\ Ghent University - Imec \And Thomas Demeester \\ Ghent University - Imec}

\begin{document}
\maketitle
\begin{abstract}
Powerful sentence encoders trained for multiple languages are on the rise.
 These systems are capable of embedding a wide range of linguistic properties into vector representations. 
 While explicit probing tasks can be used to verify the presence of specific linguistic properties, it is unclear whether the vector representations can be manipulated to indirectly steer 
 such properties. 
 For efficient learning, we investigate the use of a geometric mapping in embedding space to transform linguistic properties,  
 without any tuning of the pre-trained sentence encoder or decoder. 
 We validate our approach on three linguistic properties using a pre-trained multilingual autoencoder and 
 analyze the results 
 in both monolingual and cross-lingual settings.
\end{abstract}

\section{Introduction}

\begin{figure*}
\begin{center}
 \includegraphics[width=0.8\textwidth]{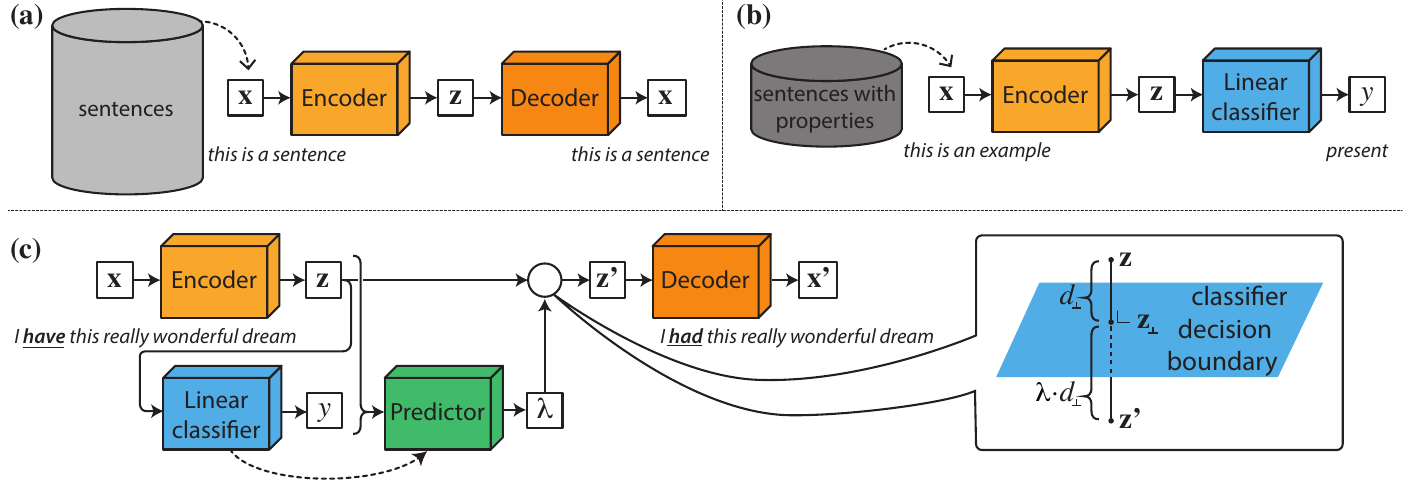}
 \caption{(a)~Pretrained autoencoder (encoder ENC, decoder DEC). (b)~linguistic property classifier $\mathcal{C}$. (c)~Geometric transformation of the sentence representation to shift $\mathbf{z}$ according to $\lambda$ beyond the decision boundary of $\mathcal{C}$, the shifted encoding $\mathbf{z}'$ is then given as input to the decoder resulting in the sentence $\mathbf{x}'$ with the transferred property.}
 \label{figure:overview}
 \end{center}
\end{figure*}

Recently,  the design of sentence encoders, monolingual \cite{kiros2015skip, conneau2017supervised} and multilingual \cite{artetxe2019massively, feng2020language} has enjoyed a lot of attention. 
 Many works have used probing tasks to investigate 
 the presence of specific linguistic properties in sentence representations \cite{adi2016fine, conneau2018you, conneau2018senteval,ravishankar2019probing,hewitt2019structural, chi2020finding}. However, it remains unclear to what extent these linguistic properties can be actually steered by manipulating the representations.
By analogy to the definition of style-transfer from \citet{li2018delete}, we refer to modifying a particular linguistic property in a given text (\eg a sentence's tense) while preserving all of the property-independent content as \emph{linguistic property transfer}.
\par
Training dedicated  models to transfer linguistic properties 
requires substantial computational effort and a lot of  training data.
Adding the ability to transform a new property may require an entire retraining of the text encoder and decoder.
 This is especially challenging for low-resource languages or when reusing or building transfer models for more than one language.  
\par 
Assuming that pre-trained autoencoders capture the linguistic properties of interest, we investigate \begin{enumerate*}[(i)]
\item whether they can be used without further tuning to efficiently transfer the properties, 
and \item whether this extends to the cross-lingual setting, when based on a multilingual pre-trained autoencoder.
\end{enumerate*}
 Our starting point is a pre-trained sentence encoder, with a corresponding decoder trained on an autoencoder objective. 
We show how a geometric transformation of pre-trained multilingual sentence embeddings can be efficiently learned on CPU for transferring specific linguistic properties.
We also experiment with cross-lingual linguistic property transfer, using a language-agnostic pre-trained encoder.

In summary, this paper presents a set of preliminary experiments on linguistic property transfer, 
and shows that there may be value in further research on manipulating distributed representations to efficiently tackle language generation tasks.

\section{Related work}
Linguistic properties usually denote the grammatical behavior of linguistic units in sentences. This contrasts with styles which concern semantic aspects of sentences such as sentiment and gender. Nevertheless, transferring linguistic properties can be situated in the broader style transfer setting.
\par
Style transfer systems can be categorized into 
\begin{enumerate*}[(i)]
    \item methods that learn \emph{disentangled} representations, in which the content is explicitly separated from the style, making the style aspect controllable and interpretable \cite{hu2017toward, shen2017style,zhao2018adversarially,fu2018style,logeswaran2018content,john2019disentangled} and
    \item methods that learn \emph{entangled} representations in which the content and style are not explicitly separated \cite{mueller2017sequence, dai2019style, liu2019revision, wang2019controllable, duan2019pre}.
\end{enumerate*}
Our approach falls under the entangled methods because
encoder-decoder systems trained on an autoencoding objective yield representations in which there is no explicit separation between content and style. Conceptually, our method is most similar to \citet{duan2019pre}, but differs in \begin{enumerate*}[(i)]
\item that it can use any existing pre-trained autoencoder as opposed to training an autoencoder from scratch on a variational objective,
\item that a  simple geometric transformation is applied on the representations instead of training a computational heavy neural transformation network, and 
\item that it generalizes to the cross-lingual setting.
\end{enumerate*}

\section{Linguistic Property Transfer}
Our 
system consists of three components:
\begin{enumerate*}[(1)]
\item a pre-trained multilingual autoencoder,
\item linear classifiers for the targeted linguistic properties 
and
\item a component 
that geometrically transforms sentence embeddings to transfer the selected properties in the dense sentence representation space.
\end{enumerate*}
These components are presented schematically in \figref{figure:overview}.
\par  We  start from a pre-trained autoencoder (\figref{figure:overview}a) that consists of an encoder ($\text{ENC}: \mathcal{X} \to \mathbb{R}^n$) which maps sentences ($\mathbf{x} \in \mathcal{X}$) to vectors ($\mathbf{z} \in \mathbb{R}^n$), and a decoder ($\text{DEC}: \mathbb{R}^n \to \mathcal{X}$) that maps the vectors  $\mathbf{z}$ back to the corresponding sentences. 
\par The second component (\figref{figure:overview}b) is a linear classifier $\mathcal{C}: \mathbb{R}^n \to \mathcal{Y}$ that takes as input a sentence encoding $\mathbf{z}$ 
and outputs a linguistic property label. We will limit our experiments to binary properties, 
\ie $\mathcal{Y} =  \lbrace 0, 1 \rbrace$. 
\par Finally, the last component (\figref{figure:overview}c), performs a geometric transformation. It
allows flipping the value of the selected linguistic property by projecting the original encoding $\mathbf{z}$ into the opposite half-space with respect to the property classifier, over an estimated distance $\lambda$. This leads to the \emph{transferred} encoding $\mathbf{z}'$, designed to be decoded into a sentence $\mathbf{x}'$ close to the original sentence, but with the transformed target property.



\par 

\par 
The three components shown in \figref{figure:overview} are further described below.
 
\subsection{Pretrained Autoencoder}
\label{subsection:pretrained}
For the pre-trained autoencoder shown in \figref{figure:overview}a, we use Language Agnostic Sentence Representations (LASER) \cite{artetxe2019massively}. LASER encodes sentences of 93 languages into a single vector space, such that semantically similar sentences in different languages have similar vectors. 
For our experiments, we leave the LASER encoder unchanged and train separate decoders for English and Dutch, by optimizing the likelihood $p(\mathbf{x}\vert\mathbf{z})$, with $\mathbf{z}=\text{ENC}(\mathbf{x})$. The decoder consists of a single-layer 1024-dimensional hidden state LSTM \cite{hochreiter1997long}.

\subsection{Linear Property Classifier}
Our approach assumes that both labels of the considered property are linearly separable in $\mathbf{z}$ space. A linear classifier $\mathcal{C}$ is trained on examples of the linguistic property.
With the coefficients $\mathbf{w}\in\mathbb{R}^n$ and bias $b\in\mathbb{R}$, its 
decision boundary is characterized by the affine hyperplane
\vspace{-1mm}
\begin{equation}\label{eqn:decision_boundary}
    \mathcal{H} = \{\mathbf{z} \in \mathbb{R}^n:  \mathbf{z} \cdot \mathbf{w} + b = 0 \}.
\end{equation}
\noindent Logistic regression was used for the results presented in this work. 

\subsection{Geometric Transformation}
The idea behind the geometric transformation is the following: a perpendicular projection from $\mathbf{z}$ onto the decision plane $\mathcal{H}$ would make the classifier $\mathcal{C}$ most uncertain about the considered attribute, with minimal changes (in Euclidean sense) to the original vector. When  removing the property information from the corresponding sentence with the opposite label, we assume it gets projected onto the same position of $\mathcal{H}$. 
As a result, the proposed geometric transformation comes down to shifting $\mathbf{z}$ in the direction perpendicular to $\mathcal{H}$, and beyond it, into the region where $\mathcal{C}$ would predict the opposite label of the property. The transformed representation $\mathbf{z}'$ is then decoded by DEC. 
The intuitive approach of simply mirroring $\mathbf{z}$ over the decision plane appears sub-optimal 
(see \secref{sec:results}). The distance into the opposite half space is therefore predicted based on the input (see \secref{subsec:CMAB}). 

The geometric shift of $\mathbf{z}$ in the direction of $\mathcal{H}$ can be derived with basic geometry, for which what follows is a brief sketch.
By construction, $\mathbf{w}$ is perpendicular to the plane described by $\mathbf{z}\cdot\mathbf{w}=0$, which in turn is parallel to $\mathcal{H}$, given \equref{eqn:decision_boundary}, such that 
$\mathbf{w}\perp\mathcal{H}$.
With that, 
the perpendicular projection $\mathbf{z}_{\perp}$ of $\mathbf{z}$ onto $\mathcal{H}$ 
can be written as
\[\mathbf{z}_{\perp} = \mathbf{z} + \beta \mathbf{w},\;\;\text{with}\;\; \beta = -\frac{\mathbf{z} \cdot \mathbf{w} + b}{\lvert \lvert \mathbf{w} \lvert \lvert^2},
\]
after substituting $\mathbf{z}_{\perp}\in\mathcal{H}$ into  \equref{eqn:decision_boundary}.
 
\noindent We finally express the transformation of $\mathbf{z}$ onto $\mathbf{z}'$ beyond $\mathcal{H}$ as 
\begin{equation}
    \mathbf{z}' = \mathbf{z}_{\perp} + \lambda(\mathbf{z}_{\perp} - \mathbf{z})
    \label{eq:zprime}
\end{equation}
where the parameter $\lambda \geq 0$ represents the distance of $\mathbf{z}'$ from $\mathcal{H}$, relative to the distance $\vert\vert \mathbf{z}_{\perp} - \mathbf{z}\vert\vert $ on the original side of the decision plane (indicated as $d_{\perp}$ in \figref{figure:overview}).

\subsection{Projection Distance Predictor}
\label{subsec:CMAB}
As mentioned above, we propose estimating
the most suitable value of $\lambda$, corresponding to how far on the other side of the decision plane $\mathbf{z}$ needs to be projected to get optimal transfer results. 
To that end, we use a contextual multi-armed bandit (CMAB) \cite{auer2002using}, a simple and efficient form of reinforcement learning with a single state, which in our setting is the sentence representation $\mathbf{z}$. 
For a new input $\mathbf{z}$, the bandit needs to select the value of $\lambda$ that best allows transferring the property with \equref{eq:zprime}, while preserving 
the content of the associated sentence $\mathbf{x}$. 
%

 The bandit method allows using a non-differentiable reward, but other choices of algorithm are possible. Our model's goal is to preserve the content of the original sentence $\mathbf{x}$ while changing its property $y$ to $y'$. Hence, our CMAB reward consists of
 \begin{enumerate*}[(i)]
     \item a linguistic property reward $r_{\text{prop}}$ and 
     \item a content-preserving reward $r_{\text{content}}$.
 \end{enumerate*}
To compute $r_{\text{prop}}$, we pass the decoded transformed sentence $\mathbf{x'}=\text{DEC}(\mathbf{z}')$ back into the encoder and use the predicted likelihood of the corresponding linear property classifier for target $y'$ as the reward:

  \begin{align*}\label{eqn:style_reward}
        r_{\text{prop}}(\mathbf{x}', y') & =  \left\{
        \begin{array}{@{} l c @{}}
            \sigma(\text{ENC}(\mathbf{x}') \cdot \mathbf{w} + b)  &  y' = 1 \\
            1 - \sigma(\text{ENC}(\mathbf{x}') \cdot \mathbf{w} + b) & y' = 0
        \end{array}\right.
 \end{align*}
 with $\sigma(.)$ the logistic function. 
\noindent For $r_{\text{content}}$, we directly optimize the BLEU-score \cite{papineni2002bleu} between the original sentence $\mathbf{x}$ and the transferred sentence $\mathbf{x}'$. Intuitively, this leads to the minimum number of changes that are required to transfer $y$ to $y'$ and thus encourages 
the content preservation between $\mathbf{x}$ and $\mathbf{x}'$:
\begin{equation*}\label{eqn:content_reward}
    r_{\text{content}}(\mathbf{x}, \mathbf{x}') = \text{BLEU}(\mathbf{x}, \mathbf{x}')
\end{equation*}
For the final reward $r(\mathbf{x}, \mathbf{x}', y')$, the harmonic mean of $r_{\text{prop}}(\mathbf{x}', y')$ and $r_{\text{content}}(\mathbf{x}, \mathbf{x}', y')$ appeared a suitable choice, encouraging the model to jointly ensure the correct target property (high $r_{\text{prop}}$) as well as preserve the sentence content (high $r_{\text{content}}$). 
\par 
We implement CMAB using the LinUCB with Disjoint Linear Models algorithm from \citet{li2010contextual}, which assumes that the expected reward obtained from choosing arm $\lambda$ is linear with respect to its input features (in our case, sentence encoding $\mathbf{z}$). For each discrete allowed value (‘arm’) for $\lambda$, LinUCB learns a separate ridge-regression model, with learnable parameters $\mathbf{A} \in \mathbb{R}^{n \times n}$ and $\mathbf{b} \in \mathbb{R}^{n}$ (for LASER $n$ is $1024$). It predicts the reward, including an upper confidence bound (UCB), for choosing that value for $\lambda$ for the given encoding $\mathbf{z}$. The hyperparameter $\alpha$ is used to control the wideness of the UCB, such that a larger $\alpha$ results in a wider UCB. Each training iteration observes a single $\mathbf{z}$ for which the arm achieving the highest potential reward (UCB) is chosen and only the parameters corresponding to its ridge-regression model are updated. Quantifying the merit of each arm for the input requires an inverse-matrix ($n \times n$) computation, 2 matrix-vector multiplications, and 2 dot products. The best arm’s parameters ($\mathbf{A}$ and $\mathbf{b}$) are then updated, requiring 1 outer-vector product. During inference, the $\lambda$ value of the best arm is used. The training and inference schemes are presented in Algorithms \ref{algo:training} and \ref{algo:inference}.

\begin{algorithm}[t!]
\SetKwInOut{Input}{input}
\Input{Exploration parameter $\alpha \in \mathbb{R}_+$ \\ $\mathcal{A} = \lbrace \lambda_1, ..., \lambda_k \rbrace$}

 \For{$(\mathbf{x}_t, y_t) \in \mathcal{D}$}{
    $\mathbf{z}_t = \text{ENC}(\mathbf{x}_t)$\\
     \For{$ \lambda \in \mathcal{A}$}{
      \If{$t = 0$} {
        $\mathbf{A}_\lambda = \mathbf{I}_{n \times n}$, 
        $\mathbf{b}_\lambda = \mathbf{0}_{n \times 1}$ \\
        }
        $\hat{\theta}_ \lambda = \mathbf{A}_\lambda^{-1} \mathbf{b}_\lambda $ \\
        $p_{t,\lambda} = \hat{\theta}_\lambda^T \mathbf{z}_t + \alpha  \sqrt{\mathbf{z}_t^T \mathbf{A}_\lambda^{-1} \mathbf{z}_t} $
        
    }
    Choose $\lambda_t = \text{argmax}_{\lambda \in \mathcal{A}}\,p_{t,\lambda}$ \\
    $\mathbf{z}_t' = \mathbf{z}_{\perp, t} + \lambda_t(\mathbf{z}_{\perp, t} - \mathbf{z}_t)$ \\
    $\mathbf{x}_t' = \text{DEC}(\mathbf{z}_t')$ \\
    $\mathbf{A}_{\lambda_t} = \mathbf{A}_{\lambda_t} + \mathbf{z}_t \mathbf{z}_t^T $ \\
    $\mathbf{b}_{\lambda_t} = \mathbf{b}_{\lambda_t} + r(\mathbf{x}_t, \mathbf{x}_t', y_{t}^{'})\mathbf{z}_t $
 }
 \caption{Training scheme, pseudocode adapted from \citet{li2010contextual}}
 \label{algo:training}
\end{algorithm}

\begin{algorithm} 
\SetKwInOut{Input}{input}
\Input{$\mathbf{A}_{\lambda}$ and $\mathbf{b}_{\lambda}$ for each arm  $\lambda \in \mathcal{A} = \lbrace \lambda_1, ..., \lambda_k
\rbrace$, \\
Sentence $\mathbf{x}$ with property label $y$
}

$\mathbf{z} = \text{ENC}(\mathbf{x})$\\
 \For{$ \lambda \in \mathcal{A}$}{
    $\hat{\theta}_ \lambda = \mathbf{A}_\lambda^{-1} \mathbf{b}_\lambda $ \\
    $p_{\lambda} = \hat{\theta}_\lambda^T \mathbf{z}$}

Choose $\lambda = \text{argmax}_{\lambda \in \mathcal{A}}\,p_{\lambda}$ \\
$\mathbf{z}' = \mathbf{z}_{\perp} + \lambda(\mathbf{z}_{\perp} - \mathbf{z})$ \\
$\mathbf{x}' = \text{DEC}(\mathbf{z}')$ \\
\Return{$\mathbf{x}'$}

 \caption{Inference scheme}
 \label{algo:inference}
\end{algorithm}

\section{Experiments}
To investigate whether linguistic properties embedded in 
representations of pre-trained encoders can be transferred without finetuning, we first apply the SentEval tool from \citet{conneau2018senteval} to LASER-embeddings (\secref{subsection:pretrained}) and identify three properties that have a strong presence. We then investigate how well our approach performs on these properties in the monolingual setting (ML), in which our CMAB model is both trained and evaluated on English sentences (\textbf{Q1}). Finally, we investigate the performance of our approach in the cross-lingual setting (CL), in which the model is trained on English but evaluated on Dutch sentences. In particular, after training on English, Dutch sentences are passed into the LASER encoder to obtain the transformed encodings $\mathbf{z}'$ which in turn are decoded by the Dutch decoder (\textbf{Q2}).

\subsection{Linguistic Properties}
\begin{table}
\label{table:probing}
\small
\centering
\begin{tabular}{ll} 
 \toprule
 \multicolumn{2}{c}{Probing Task: Accuracy (\%)} \\ 
 \midrule
  Length: 74.09  & \textbf{Tense}: \textbf{89.1} \\ 
  BigramShift: 68.06 & CoordinateInversion: 67.82\\ 
  OddManOut: 50.80 &  Depth: 39.2 \\  
  TopConstituents: 39.2 & \textbf{SubjNumber}: \textbf{90.69} \\ 
  \textbf{ObjNumber}: \textbf{88.72}  &  \\ 
\bottomrule
\end{tabular}
\caption{Results of LASER-embeddings on the probing tasks of \citet{conneau2018senteval}. In our experiments, we transfer the properties denoted in bold.}
\label{table:probing-tasks}. 
\end{table}

Table \ref{table:probing-tasks} shows the results of LASER-embeddings on the probing tasks from \citet{conneau2018senteval}. The high accuracies for the properties shown in bold, indicate that LASER encodes them well. 
In our experiments, we transfer
\begin{enumerate*}[(i)]
\item the \textbf{Tense} of the main verb which is either in the present or past,

\item \textbf{ObjNum}, representing the number (singular or plural) of the main clause's direct object and
\item \textbf{SubjNum}, which is the number (singular or plural) of the subject of the main clause. 
\end{enumerate*}

\subsection{Implementation and Training Data}
As discussed in \secref{subsection:pretrained}, we use LASER's encoder and train two decoders on it with around 20M English and Dutch OpenSubtitles sentences \cite{tiedemann2012parallel,lison2019open}.
For each property, we train a binary logistic regression model on CPU using SentEval data, through stratified 5-fold cross-validation. 
We found that training the CMAB-models on SentEval led to worse results than training on OpenSubtitles. We hypothesize that this is due to a mismatch between the SentEval  -and OpenSubtitles sentences on which the decoders were trained. We therefore trained, on CPU, the CMAB-models using 2500 English OpenSubtitles sentences with (noisy) property labels predicted by the SentEval classifiers.
Across all experiments, we use the discrete set $\lbrace 1, 1.5, \ldots, 7 \rbrace$ as possible values for $\lambda$ (`arms' of the CMAB algorithms)
and set the CMAB exploration parameter $\alpha$ 
to 4.

\subsection{Evaluation} 
We randomly selected OpenSubtitles sentences (not seen during decoder training), and for those with any of the target properties present, annotated the corresponding sentence with the flipped property.  
As such, 100 test-pairs $(\mathbf{x}, \mathbf{x}')$ were obtained for each property.
We report human evaluation metrics:
\begin{enumerate*}[(i)]
\item the percentage of transferred sentences that have the correct property (`Label' accuracy), and
\item the percentage of transferred sentences that have the correct property  \emph{and} preserve the content (`All' accuracy).
\end{enumerate*}
We also include the BLEU-score between the transferred sentence and the gold target $\mathbf{x}'$.

\subsection{Results}
\label{sec:results}
To answer (\textbf{Q1}), we refer to the first three rows of Table \ref{table:Q1}. Our approach switches properties in roughly half of the cases (label accuracy).
However, fewer cases occur in which both the property is transferred and content is preserved. The last three rows of Table \ref{table:Q1} display the metrics in the cross-lingual setting in which we notice similar results as in the previous setting (\textbf{Q2}). 
The results are encouraging, although we expect further improvements from more complex transformation approaches.
Table \ref{table:baseline} shows, for \textbf{Tense}\textsubscript{ML}, a comparison of our CMAB approach against a baseline, that mirrors each $\mathbf{z}$ over the decision boundary i.e, $\lambda=1$. We find that the CMAB-approach outperforms that baseline for all metrics. Moreover, Table \ref{table:arm-distributions} shows the distribution of the predicted arms on the test sets in the monolingual and cross-lingual settings, indicating that choosing the optimal value for $\lambda$ is input-dependent. As an illustration, Table \ref{table:examples} lists a few examples, picked randomly from among those test items with successful label transformation and content preservation.


\begin{table}
\small
\centering
\begin{tabular}{lcccc} 
 \toprule
 & Property & \multicolumn{1}{c}{Label (\%)} & \multicolumn{1}{c}{All (\%)} & \multicolumn{1}{c}{BLEU}  \\ 
 \midrule
  Mono-&\textbf{Tense}\textsubscript{ML} & 61 & 47 &  54.9  \\  
  lingual&\textbf{ObjNum}\textsubscript{ML} & 44 & 29 & 39.0  \\ 
  &\textbf{SubjNum}\textsubscript{ML} & 48 & 34 & 36.3   \\  
  \midrule
  Cross-&\textbf{Tense}\textsubscript{CL} & 51   & 41 & 49.9  \\  
  lingual&\textbf{ObjNum}\textsubscript{CL} & 49 & 43   & 49.0  \\  
  &\textbf{SubjNum}\textsubscript{CL} & 56 & 33 & 32.6   \\  
\bottomrule
\end{tabular}
\caption{Human label accuracy (`Label') and accuracy of both label and content (`All'), and BLEU-scores of our CMAB-approach (monolingual and cross-lingual).}
\label{table:Q1}. 
\end{table}

\section{Conclusion and Future Work}
\label{sec:conclusion}
We have introduced a simple and efficient geometric method to transfer linguistic properties which has been evaluated on three properties in both monolingual and cross-lingual settings. While there is room for improvement, our preliminary results indicate that it can allow pre-trained autoencoders to transfer linguistic properties without additional tuning, such that there is no need to train dedicated transfer systems. This potentially makes learning faster and better scalable than with existing methods.
For future work, we aim at extending our method to transformer-based encoders (monolingual and cross-lingual), and will consider additional linguistic as well as more style-oriented properties.
\begin{table}
\small
\centering
\begin{tabular}{cccc} 
 \toprule
  Model & Label (\%) &  All (\%) & BLEU  \\ 
  \midrule
  Baseline & 59 & 28 & 53.1 \\
  CMAB & \textbf{61} & \textbf{47} & \textbf{54.9} \\
\bottomrule
\end{tabular}
\caption{Comparison of the baseline ($\lambda=1$) and the CMAB-approach for \textbf{Tense}\textsubscript{ML}.}
\label{table:baseline}. 
\end{table}

\begin{table}
\small
\centering
\begin{tabular}{ccccccc} 
 \toprule
 $\lambda$ & \multicolumn{1}{c}{\textbf{Tense}\textsubscript{ML(CL)}} & \multicolumn{1}{c}{ \textbf{SubjNum}\textsubscript{ML(CL)} } & \multicolumn{1}{c}{\textbf{ObjNum}\textsubscript{ML(CL)}} \\
 \midrule
 $\mathbf{1}$   & 2.5(5)     & \xmark     & \xmark  \\
 $\mathbf{1.5}$ & 23.5(31.5) & \xmark     & \xmark  \\
 $\mathbf{2}$   & 43.5(34.5) & \xmark     & \xmark  \\
 $\mathbf{2.5}$ & 14(19)     & \xmark    & 13(14)  \\
 $\mathbf{3}$   & 15(7)      & 3(1)     & \xmark   \\
 $\mathbf{3.5}$ & \xmark    & \xmark     & \xmark  \\
 $\mathbf{4}$   & 1.5(3)     & 21(\xmark)    & \xmark  \\
 $\mathbf{4.5}$ & \xmark    & \xmark(29.5)  & \xmark  \\
 $\mathbf{5}$   & \xmark    & 5.5(6.5) & 1.5(5.5)  \\
 $\mathbf{5.5}$ & \xmark    & 21(26)   & \xmark  \\
 $\mathbf{6}$   & \xmark    & 24(11.5) & 0.5(50)  \\
 $\mathbf{6.5}$ & \xmark    & 8.5(13)  & 22(15)  \\
 $\mathbf{7}$   & \xmark    & 17(12.5) & 63(15.5) \\
\bottomrule
\end{tabular}
\caption{Distributions of the predicted projection distances of the CMAB for the different test sets expressed as a percentage (monolingual and cross-lingual).}
\label{table:arm-distributions}. 
\end{table}

\begin{table}[t!]
\small
\centering
\begin{tabular}{ll} 
 \toprule
 & \multicolumn{1}{l}{\textbf{Tense} (present$\to$past)} \\
 \midrule
  Mono-& i ask many people here . \\
  lingual& i \textbf{asked} many people here . \\ \midrule
  Cross-& ik kijk naar een oude film van m ' n moeder . \\
  lingual& ik \textbf{bekeek} een oude film van mijn moeder . \\
  \toprule
  & \multicolumn{1}{l}{\textbf{ObjNum} (singular$\to$plural)} \\
  \bottomrule
    Mono-& i could tell you some story . \\
    lingual& i could tell you some \textbf{stories} . \\ \midrule
    Cross-& we hebben een beter bondgenoot nodig . \\
    lingual& we hebben \textbf{betere bondgenoten} nodig . \\
    
  \toprule
    & \multicolumn{1}{l}{\textbf{SubjNum} (plural$\to$singular)} \\
    \bottomrule
    Mono-& families agreed to keep it quiet . \\
    lingual& \textbf{a family} agreed to keep it quiet . \\ \midrule
    Cross-& monsters gaan ons opeten . \\
    lingual&\textbf{het monster gaat} ons opeten . \\
\bottomrule
\end{tabular}
\caption{Linguistic property transfer examples of the proposed system in both monolingual and cross-lingual settings}
\label{table:examples}
\end{table}

\vspace{-8pt}
\section*{Acknowledgments}
\noindent This work was funded by the Flemish Government (VLAIO), Baekeland project-HBC.2019.2221. 

\bibliography{anthology,custom}
\bibliographystyle{acl_natbib}

\end{document}